\documentclass{article}

\usepackage{arxiv}

\usepackage[utf8]{inputenc} % allow utf-8 input
\usepackage[T1]{fontenc}    % use 8-bit T1 fonts
\usepackage{hyperref}       % hyperlinks
\usepackage{url}            % simple URL typesetting
\usepackage{booktabs}       % professional-quality tables
\usepackage{amsfonts}       % blackboard math symbols
\usepackage{nicefrac}       % compact symbols for 1/2, etc.
\usepackage{microtype}      % microtypography
\usepackage{lipsum}		% Can be removed after putting your text content
\usepackage{graphicx}
\usepackage{natbib}
\usepackage{doi}

\usepackage{amsmath}
\usepackage{xcolor}
\usepackage{textcomp}
\usepackage{listings}

\title{Hebbian learning with gradients: Hebbian convolutional neural networks with modern deep learning frameworks}

\date{} 					% Or removing it

\author{\href{https://orcid.org/0000-0002-7897-4492}{\includegraphics[scale=0.08]{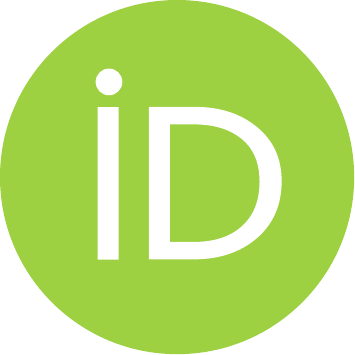}\hspace{1mm}Thomas Miconi}
\\
	ML Collective\\
	\texttt{thomas.miconi@gmail.com} \\
}

\renewcommand{\headeright}{}
\renewcommand{\undertitle}{}
\renewcommand{\shorttitle}{Multi-layer Hebbian networks}

\begin{document}
\maketitle

\begin{abstract}

Deep learning networks generally use non-biological learning methods. By contrast, networks based on more biologically plausible learning, such as Hebbian learning, show comparatively poor performance and difficulties of implementation. Here
we show that Hebbian learning in hierarchical, convolutional neural networks can be implemented almost trivially with modern deep learning frameworks, by using specific losses whose gradients produce exactly the desired Hebbian updates. We provide expressions whose gradients exactly implement a plain Hebbian rule ($\Delta{w} \propto xy$), Grossberg's instar rule ($\Delta{w} \propto y(x-w) $), and Oja's rule ($\Delta{w} \propto y(x-yw)$).
As an application, we build Hebbian convolutional multi-layer networks for object recognition. We observe that higher layers of such networks tend to learn large, simple features (Gabor-like filters and blobs), explaining the previously reported decrease in decoding performance over successive layers. To combat this tendency, we introduce  interventions (denser activations with sparse plasticity, pruning of connections between layers) which result in sparser learned features, massively increase performance, and allow information to increase over successive layers. We hypothesize that  more advanced techniques (dynamic stimuli, trace learning, feedback connections, etc.), together with the massive computational boost offered by modern deep learning frameworks, could greatly improve the performance and biological relevance of multi-layer Hebbian networks.
\end{abstract}

% keywords can be removed
% \keywords{First keyword \and Second keyword \and More}

\section{Introduction}

Recent advances in deep learning have greatly improved the state of the art in visual object recognition. Deep learning networks take inspiration from features of the ventral visual cortex (including a hierarchical organization in successive layers and a topological organization of connections). Their internal activity shows similarities, as well as differences, with neural responses to natural images \citep{yamins2016using}. One important difference is that deep learning networks are usually trained by gradient descent and backpropagation. By contrast, learning in the visual cortex is generally thought to occur primarily by local learning (with some influences from both downstream and upstream activity), presumably according to the Hebbian principle of reinforcing connections that cause a cell to fire. Investigating Hebbian learning in large hierarchical networks could therefore provide a more relevant model of the visual system, as well as offering new methods for unsupervised and self-supervised (label-free) learning.

There is a long history of research in hierarchical Hebbian networks. The Neocognitron (an early hierarchical model for translation- and scaling-tolerant object recognition) used an effectively Hebbian learning rule \citep{fukushima1982neocognitron}. The HMAX model used an ``imprinting'' learning method to approximate Hebbian learning in a large network in a computationally tractable manner \citep{serre2007feedforward}. The VISNet model also studied various forms of Hebbian learning with temporal dynamics to learn invariant representations \citep{wallis1997invariant,rolls2000model}. While these models and others provided rich insight, their performance remained far from current deep learning techniques. Some of this difference may result from the massive computational advances that supported (and were prompted by) the deep learning revolution, including both specialized hardware \citep{pinto2009high,krizhevsky2012imagenet} and software frameworks to interact with this hardware in an intuitive manner \citep{paszke2019pytorch,abadi2015tensorflow}.

Another motivation for investigating Hebbian networks is that Hebbian learning can be generalized to implement many forms of learning, including sparse coding \citep{brito2016nonlinear}, K-means \citep{hu2014modeling} and similarity-matching \citep{bahroun2017online}. However, pure Hebbian multi-layer networks suffer from poor performance. Notably, a consistent result is that decoding performance \emph{decreases} over successive layers, suggesting loss of information \citep{amato2019hebbian,bahroun2017building}. This is a particularly disconcerting finding, since the purpose of hierarchical networks is precisely to aggregate information and produce more exploitable representations over successive layers \cite{dicarlo2007untangling,serre2007feedforward}.

Recently several authors have proposed using modern deep learning framework for Hebbian learning in multi-layer convolutional networks \citep{amato2019hebbian,talloen2020pytorch}. These projects rely on intricate hand-crafted machinery to implement Hebbian learning. While \citet{talloen2020pytorch} expose this machinery as a remarkable reusable framework, this necessarily constrains possible usage to whatever abilities the framework provides, in addition to burdening users with the need to learn the framework itself. 

Here we show that Hebbian learning can be implemented very easily with modern deep learning frameworks, by using surrogate losses whose gradients turn out to produce exactly the desired Hebbian updates. As a result, we use the frameworks as they were intended to be used, with little additional burden, freeing us to concentrate on actual network design. Importantly, the actual outputs of the network are arbitrary and need not be differentiable, since the surrogate losses are only used to set up the computational graph. We provide expressions for losses whose gradients exactly implement several Hebbian rules, namely: plain Hebbian ($\Delta{w} \propto x y$), Grossberg's instar rule ($\Delta{w} \propto y (x-w) $), and Oja's rule ($\Delta{w} \propto y (x-yw)$).

As an application, we build simple convolutional multi-layer networks for object recognition, which we test on the CIFAR10 dataset. We observe that the higher layers of such networks do not spontaneously learn complex shapes. Rather, they simply learn large, simple features (Gabor-like oriented edge detectors and blobs), by  combining lower-level features appropriately. This explains the reduction in decoding performance over successive layers, despite a higher number of filters, as was observed in previous work \citep{amato2019hebbian,bahroun2017building,bahroun2017online}. To combat this tendency, we introduce several interventions (``triangle'' method for computing activations \citep{coates2011analysis} and massive pruning of connections between layers) which both prevent the formation of high-level Gabors and massively increase higher-level performance, allowing higher layers to produce more informative representations than the first layer. We hypothesize that incorporating more advanced features of previous models (dynamic stimuli with trace learning, feedback connections, etc.), together with the massive computational boost offered by modern deep learning frameworks, could greatly improve the performance and biological relevance of multi-layer Hebbian networks.

\section{Methods}

\subsection{Hebbian learning with deep learning frameworks}

Modern deep learning frameworks implement automatic differentiation with transparent access to GPU computation, batched processing, and convolutional filtering. As such, they are ideally suited to compute batch gradients over multi-layer convolutional networks \citep{paszke2019pytorch,abadi2015tensorflow}.

Implementing Hebbian learning poses no difficulty with fully-connected layers, or fully recurrent networks, because the input and output are simple vectors (or matrices when considering the batch dimension). As a result, computing the Hebbian weight update (based on a product of inputs and outputs) requires little more than an outer vector product. No special machinery is needed \citep{miconi2018differentiable}.

Things are not so simple with convolutional networks, because in this case the inputs to the weights are composed of \emph{overlapping} patches within the previous layer's output. As such, we can not simply grab the inputs and outputs and perform a straightforward outer product. We propose two methods, not necessarily exclusive, to implement such computations with deep learning frameworks.

\subsubsection{Method 1: Hebbian updates as gradients of surrogate losses}

By far the simplest method is to compute a \emph{surrogate loss}, the gradient of which just happens to be the desired Hebbian weight update. 

We first compute the ``real'' output of that layer, which is arbitrary and can involve any computations, differentiable or not (in this work we mostly use a winner-take-all competition, which is not straightforwardly differentiable). Then, we compute a \emph{surrogate loss} for this layer, designed specifically to produce the desired update as a gradient (see Appendix \ref{sec:hebb} for equations and code samples). After this surrogate loss is computed, setting in place the appropriate computational graph, we overwrite the values of this surrogate output with the ``real'' output values previously computed, without impacting the computational graph (because the gradient of the surrogate loss must be evaluated at the ``real'' value of the output). Then we simply invoke the framework's backward pass function to compute and apply the desired weight updates (see Appendix \ref{sec:hebb} for details).

At first sight this seems to require two separate forward passes: one to compute the real output, and one to compute the surrogate loss. However, this is not the case, because the computationally intensive element (namely, the linear convolution by the weight filters) is common to both real and surrogate output, and thus only needs to be computed once. The only additional computation is the surrogate loss function, which is generally negligible in comparison to the convolution itself. 

Equations and code samples for plain Hebb, Instar and Oja's rule are provided in Appendix \ref{sec:hebb}. All code for the experiments described in this paper is available at
\url{https://github.com/ThomasMiconi/HebbianCNNPyTorch}.

\subsubsection{Method 2: Unfolding the input with fixed convolutions}

Alternatively, we can eliminate overlapping input patches by separating the weight convolution into two successive convolutions: a fixed, binary convolution that rearranges every input patch into a single column vector, followed by a  $1\times1$ convolution that contains the actual weights. More precisely, suppose our original convolution has input size $h \times w \times n_i$ (where $h$ and $w$ are the height and width of the convolutional filter, and $n_i$ is the number of channels in the input), with $n_o$ output channels. Then, we can first pass the input through a fixed convolution of input size $h \times w \times n_i$ with $h w n_i$ output channels, with a fixed weight vector set to 1 for the weights that links input $x, y, i$ to output $x y i$ (where $x$, $y$ and $i$ run from 1 to $h$, $w$ and $n_i$ respectively) and 0 everywhere else. This rearranges (and duplicates) the values of each input patch of the original convolution into single, non-overlapping column vectors. Afterwards we can apply the actual weights of the original convolution with a simple $1\times1$ convolution, which can be performed by a simple tensor product with appropriate broadcasting if necessary.

This somewhat clunkier method does require two convolutional steps, as well as additional memory usage. However, it also provides finer-grained control. For example, it makes it easy to separately normalize and whiten each input patch of the convolution separately, which is commonly done in computer vision models \citep{coates2011analysis,olshausen1997sparse}, but not easy to do with a normal convolution. In addition, it makes it possible to compute the Hebbian weight update by hand, since the inputs of each weight application are neatly separated into non-overlapping vectors. This may be of interest if one seeks more complex updates that are not easily reducible to a gradient. Conversely, note that this method and the previous one are not mutually incompatible: it is possible to use Method 2 to provide individual patch normalization (and whitening, see below), and then Method 1 to compute the actual weight updates.

All results reported in this work use Method 1 exclusively. In our experiments, individual patch normalization did not seem to improve performance (though individual patch whitening in the first layer did improve performance somewhat over the whole-image whitening method described below), and we are not studying any Hebbian rule beyond the ones mentioned above, making the added complexity of Method 2 superfluous. 

\subsection{Whitening}

Natural images exhibit considerable correlations between pixels. It is generally desirable to remove this correlation in order to make the actual signal easier to extract \citep{olshausen1997sparse,hyvarinen2000independent,coates2011analysis,bahroun2017online}. In grayscale images, this correlation mostly takes the form of spatial autocorrelation, resulting in much higher energy at low frequencies; as a result, it is usually eliminated by a simple band-pass filter (in the spatial domain, such filters tend to resemble the center-surround structure of retinal receptive fields) \citep{olshausen1997sparse}. In color images, however, the correlations are more complex and involve inter-channel interactions. As a result, it is customary to use a more complex form of whitening, such as the so-called ZCA method which makes the covariance matrix of the transformed data equal to the identity matrix  \citep{bell1997independent,krizheksy09learning,coates2011analysis,bahroun2017building,hyvarinen2000independent}. We first normalize each image vector $\mathbf{x}$ (i.e. subtract the mean and divide by the standard deviation\footnote{In theory, the mean and standard deviations are to be taken over the dataset. In practice, we found that taking them over each individual image works well, and indeed better than taking them over the batch dimension.}), then multiply it by the matrix 
$\mathbf{ED}^{-1/2}\mathbf{E}^\text{T}$, where $\mathbf{E}$ is the matrix of the eigenvectors of the covariance matrix of the $\mathbf{x}$ vectors and $\mathbf{D}$ the diagonal matrix of its eigenvalues (to which we add a regularization constant of $10^{-3}$ to avoid instabilities). Note that the spatial domain visualization of this filter also has a center-surround structure, though one which depends on the preferred color of the center and surround neurons (see e.g. Figure 1.3 in \citep{krizheksy09learning}), suggesting a plausible biological implementation. In this work we first compute the ZCA matrix over the entire training set, then apply it to each input image during training and testing.

\subsection{Network architecture}

All experiments use the CIFAR10 dataset \cite{krizheksy09learning}, which contains 60000 color images of size $32 \times 32$ pixels (50000 for training and 10000 for testing). Each image is centered and whitened as explained above before being fed to the network.

Our network uses three convolutional layers, with kernel size $5 \times 5$ in the first layer and $3 \times 3$ in the others. Before each layer, we normalize each element in the batch by subtracting its mean and dividing it by its standard deviation (over all positions and channels). 

Our basic network applies a winner-take-all selection (WTA) to each layer's outputs: for each position, we only preserve the channel with the highest activation and set all other ones to 0 (the winning channel output is set to 1; while not strictly necessary, this binarizing makes learning more stable). This choice parallels previous work \citep{fukushima1982neocognitron,masquelier2007learning} and is influenced by \citet{hu2014modeling}'s proof that Hebbian plasticity with WTA in the outputs is essentially equivalent to K-means clustering, which is known to produce highly efficient and biologically plausible receptive fields \citep{coates2011analysis,hu2014modeling}. Later we describe different forms of selection that improve performance considerably over this baseline.

After each layer a $2 \times 2$ average pooling and downsampling is applied (note that plasticity occurs before pooling, so pooling at a layer has no effect on the learning in the same layer). We use relatively small networks with 100, 196 and 400 filters at layers 1, 2 and 3 respectively (196 is the closest perfect square to 200; we prefer square numbers of filters for presentation purposes). With these settings, receptive fields at layer 1, 2 and 3 have size $5\times5$, $10\times10$ and $20\times20$ pixels, respectively. The final output of the network has dimensions $400 \times 2 \times 2$. We train the network over 20 epochs, each containing the whole training set; we verified that this duration was sufficient for stabilizing the learned weights.

In addition, we use adaptive thresholds to ensure similar firing rates across neurons: after computing the total input to each neuron, but before applying WTA, we add a bias term to each neuron's activation that is adaptively modulated over time to ensure the firing rate of the neuron remains close to $1/N$, where N is the number of neurons in the layer (when using $k$-WTA, the target rate for each neuron is instead $k$/N). These thresholds are not strictly necessary, but help prevent the appearance of ``dead'' cells that never fire at all (though we note that such cells are in themselves an object of study \citep{shoham2006silent,thorpe2011grandmother}).

We also keep the weight vectors of all neurons normalized to Euclidean norm 1. While Oja's rule can theoretically constrain the norms of weight vector to one under some conditions, in practice we found that it did not fully stabilize weights in our settings (presumably due to interference from nonlinearities and adaptive thresholds).

In all experiments, we use Grossberg's Instar rule \citep{vasilkoski2011review}, also used in previous work \citep{hu2014modeling,amato2019hebbian}. This is implemented by the surrogate gradient method described above. In practice, we found that with weights constrained to norm 1, the choice of plain Hebbian, Instar or Oja's rule made little difference in the results.

\subsection{Decoding and evaluation}

We assess network performance with a simple linear decoder: we freeze weights, then feed the entire training set to the network and regress the one-hot label vectors over the network outputs. We then apply these regression weights to the network's response to test set images, and pick the category with the largest output for each image. This simple regression classifier is orders of magnitude faster than the linear SVM used in other work \citep{coates2011analysis,bahroun2017building}, at only a minor performance cost.

To evaluate the information contained in each layer below the top one, we use the ``quadrants'' method introduced by \citet{coates2011analysis}. We divide the layer's output spatially in four equal quadrants, average the value of each channel across all positions within each quadrants, and concatenate the results. For layer 1, this results in a vector of size 400 (100 filters times 4 quadrants), to which we apply the linear decoding procedure described above. We note that the quadrants method is an extremely powerful baseline, reaching a test accuracy of $43\%$ on CIFAR10 from the output of the first layer of an \emph{untrained} network, i.e. with randomly initialized weights (see Table \ref{table:results}). Note that for the last layer, because the network output has spatial dimensions $2\times2$, the quadrants method is identical to simply using the output of the network as it is. 

\begin{figure}
    \centering
    \includegraphics[scale=.5]{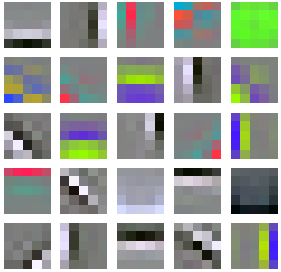}
    \hspace{1pc}
    \includegraphics[scale=.5]{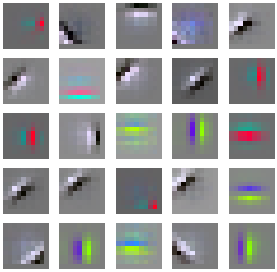}
    \hspace{1pc}
    \includegraphics[scale=.5]{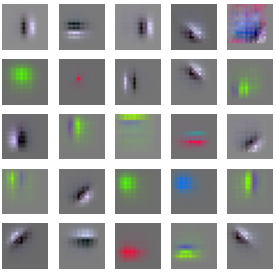}
    \caption{Example receptive fields from layers 1 (left), 2 (center) and 3 (right) in a standard Hebbian network with WTA competition. Receptive fields have size $5\times5$ (layer 1), $10\times10$ (layer 2) and $20\times20$ (layer 3). The full sets of receptive fields for each layer are shown in Figure \ref{fig:RFs_pm0_full}.}
    \label{fig:RFs_pm0}
\end{figure}

The receptive fields of layer 1, which take input directly from the RGB representation of input images, can be visualized by simply using the weights as image pixels (since weights can be negative, we normalize to the $[0,1]$ range, ensuring that all receptive fields use a common value for weight 0). The receptive fields of higher levels are reconstructed as the weighted sums of previous-layer receptive fields at the appropriate location. This reconstruction is necessarily approximate, due to the pooling operation after each layer. Figures \ref{fig:RFs_pm0} and \ref{fig:RFs_standard} show example receptive fields after training (full sets of learned receptive fields are shown in Figures \ref{fig:RFs_pm0_full} and \ref{fig:RFs_standard_full}).

\section{Results}

\subsection{Unconstrained hierarchical Hebbian learning builds simple higher-level features}

We observe poor performance on CIFAR10, reaching a test accuracy of $34.5\%\pm0.4$ (mean and standard deviation over 10 runs). As previously reported \citep{bahroun2017building,bahroun2017online,amato2019hebbian}, we observe that object information actually \emph{decreases} across successive layers, despite the increasing number of channels. In particular, the quadrant methods applied to layers 1 and 2 result in test accuracies of $49.4\%\pm0.4$ and $45.3\%\pm0.6$, respectively (see Table \ref{table:results} for a summary of results).  

Inspection of learned receptive fields provides an explanation for this poor performance. As expected from previous work \citep{krizheksy09learning,bahroun2017building}, layer 1 receptive fields include a mixture of grayscale Gabor-like features, color-opponent cells (cyan-red, blue-orange and green-purple), and flat fields of white, black, red, green and blue (this outcome, including the specific colors, was extremely reproducible across many parameter and design choices). However, unlike surmised in traditional models of the visual system \citep{serre2007feedforward,wallis1997invariant}, the higher layers do not learn increasingly complex, shape-selective combinations of lower-level features. Instead, they simply learn larger Gabor-like features and poorly localized blobs. Notably, the layers do not simply perform an identity mapping, merely transmitting specific individual lower-level channels (as observed in the single-column case by \citet{olshausen1997sparse}); instead, they precisely combine lower-level features to form higher-level, larger oriented edge detectors and blobs. As a result, the information in higher layers is essentially a less spatially precise version of that provided by layer 1, making the loss of decoding performance unsurprising. 

Varying the number of winners in each layer (that is, using $k$-WTA instead of WTA, for various values of $k$ ranging from 1 to hundreds) had only minimal impact on performance, and simply resulted in larger, more blurry receptive fields. 

\subsection{Dense responses, pruned networks and dispersed weights improve performance}

We introduce two manipulations which, put together, considerably improve performance and produce very different receptive fields.

\begin{enumerate}
    \item We relax the competition between neurons for the purpose of computing activations, but \emph{not} for the purpose of applying plasticity. Activations are now computed by the ``triangle'' method introduced by \citet{coates2011analysis}: at each position, we subtract the mean activation (across all channels at this position) from all channels, and rectify negative values to 0 (before pooling and downsampling). This results in less sparse, but still competitive responses: for each stimulus, on average, about half the cells at any given column have non-zero response. Importantly, this determines the responses transmitted to the next layer, but \emph{not} the plasticity: the output used for plasticity at each position is still determined by binary WTA, as before (that is, only the most active neuron in any column undergoes plasticity). This differentiation between selection for activation and for plasticity  considerably improved performance over alternative choices (e.g. WTA or $k$-WTA for both responses and plasticity, or triangle method for both).

\item We massively prune the connectivity of the network, such that only 1\% of all possible connections can be nonzero. This is done by generating a random binary mask for each weight vector at the beginning of the experiment, containing 99\% of zeros and 1\% of ones, and multiplying the weight vectors by their mask after every weight modification (we still normalize the pruned weights to norm 1). 

\end{enumerate}

Note that these methods are only applied to layers 2 and above, since layer 1 is supposed to take inputs directly from the ``retinal'' image input. 

These manipulations, put together, greatly improve performance.  With these alterations, the model reaches a test accuracy of  $64.55\%\pm0.5$ (mean and standard deviation over 10 runs). This is well above the decoding accuracy from the quadrants method on layer 1 ($58.56\%\pm0.3$), and similar to the same from layer 2 ($64.2\pm0.3$), showing that information now increases or persists over successive layers (see Table \ref{table:results} for a summary of results).

\begin{figure}
    \centering
    \includegraphics[scale=.5]{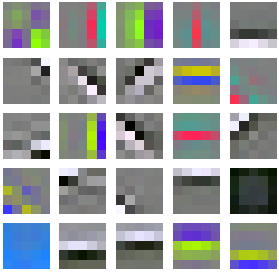}
    \hspace{1pc}
    \includegraphics[scale=.5]{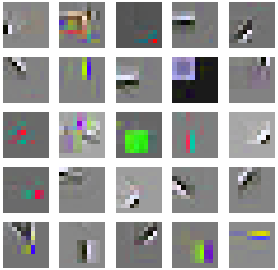}
    \hspace{1pc}
    \includegraphics[scale=.5]{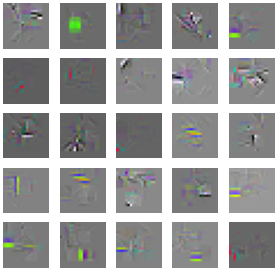}
    \caption{Example receptive fields from layers 1 (left), 2 (center) and 3 (right) in a Hebbian network with dense activations and pruned connections (plasticity still occurs through WTA competition).  The full sets of receptive fields for each layer are shown in Figure \ref{fig:RFs_standard_full}. }
    \label{fig:RFs_standard}
\end{figure}

\section{Discussion}

\begin{table}
	\caption{Test accuracy on CIFAR10, for the default network (unpruned weights, WTA competition) and the pruned network with triangle method, before training (i.e. at initialization) and after Hebbian training on 20 epochs over the training dataset.}
	\centering
	\begin{tabular}{cccc}
 		\toprule
% 		\multicolumn{2}{c}{Part}                   \\
% 		\cmidrule(r){1-2}
%		Name     & Description     & Size ($\mu$m) \\
        & L1 quadrants & L2 quadrants & Final output \\
		\midrule
		Default network, untrained $w$& $43.3\%\pm0.7$  & $21.05\%\pm0.5$ & $10.56\%\pm0.4$ \\
		Default network, trained $w$ & $49.94\%\pm0.4$ &  $45.3\%\pm0.6 $ & $34.5\%\pm0.4  $   \\
		Network with triangle method \& pruning, untrained $w$    & $52.38\%\pm0.4$ & $57.17\%\pm0.3$ & $56.3\%\pm0.3$  \\
		Network with triangle method \& pruning, trained $w$ & $58.56\%\pm0.3$ & $64.2\%\pm0.4$ & $64.55\%\pm0.4$  \\
		\bottomrule
	\end{tabular}
	\label{table:results}
\end{table}

Transparent use of deep learning frameworks for multi-layer Hebbian learning opens the door to many avenues of exploration. In this work, we showed that unconstrained Hebbian learning tends to settle on simple, information-poor features, explaining the paradoxical reduction in information and decoding performance over successive layers that had been previously reported \citep{bahroun2017building,amato2019hebbian}. We countered this with a more permissive selection in activations (but not for plasticity), and a pruning of synaptic trees. However, these only scratch the surface of possibilities. Further possible directions of research include:

\begin{enumerate}

    \item \emph{Dynamic stimuli and temporal learning rules:} There is a long history of using the temporal structure of visual stimuli to extract transformation-invariant representations \citep{foldiak1991learning,wallis1997invariant,masquelier2007learning}. In this work we have only considered static stimuli, but it would be highly desirable to exploit the wealth of temporal information available in natural stimuli.
    \item \emph{Decorrelation in higher layers:} Here we use ZCA decorrelation in the first layer, and rely on WTA to prevent highly correlated firing. More aggressive forms of decorrelation in the upper layers might prove beneficial. We experimented with channel-wise lateral inhibition and online ZCA, but neither improved performance. More elaborate (or more biologically inspired) scheme might produce different results.
    \item \emph{Feedback connections:} in the visual system, feedback connections carry information from upper to lower layers. The computational role of these connections has been debated \citep{lee2003hierarchical,hupe1998cortical,rao1999predictive}. Recent computational work suggests that appropriately configured feedback can greatly improve learning \citep{payeur2021burst,lindsey2020learning}. How to incorporate feedback connections into large networks remains an area of active research, which could greatly benefit from the computational power of deep learning frameworks.
\end{enumerate}

We note that, although our interventions greatly improved the performance of the network, the learned features (see Figure \ref{fig:RFs_standard}) are difficult to interpret. They are also quite different from the organized, shape-selective features posited by hierarchical models of the visual cortex. In these models, lower layers are expected to respond to simple conjunctions of edges and angles, while intermediate layers respond to middle-scale components such as curvature, and the higher layers develop selectivity for individual objects and faces across multiple poses. This raises the question: how can biologically plausible hierarchical networks develop selectivities to specific, increasingly complex stimuli, elaborating their selectivities across successive layers in a way that matches the increasing specificity of visual cortex? We suggest that the computational power of modern deep learning frameworks is likely to prove extremely useful for such investigations.

In another direction, several meta-learning approaches seek to discover new learning rules for particular domains \cite{bengio1997optimization,risi2010indirectly,najarro2020meta}. The method described above could conceivably be harnessed to search over possible loss expressions (as functions of $x, y, w$ or any other variable accessible to the framework), seeking losses whose gradients would produce optimal learning, whether in unsupervised, self-supervised, or supervised learning. 

In conclusion, we showed that modern deep learning frameworks are a powerful tool for exploring Hebbian learning in multi-layer convolutional networks. We believe these frameworks offer exciting opportunities for further research, both in terms of performance for unsupervised and self-supervised learning, and of biological relevance and insight for models of the visual cortex.

\appendix
\section*{Appendix}

\renewcommand\thefigure{\thesection.\arabic{figure}}   

\setcounter{figure}{0}  

\section{Hebbian updates with gradients}
\label{sec:hebb}

All code for the experiments described above is available at %\href{https://github.com/ThomasMiconi/HebbianCNNPyTorch}
\url{https://github.com/ThomasMiconi/HebbianCNNPyTorch}.

We want to build a surrogate loss function, such that the gradient of this loss is equal to the desired Hebbian update. To do this, we note that the loss $\mathcal{L}$ is a function of the network output $y$, and thus its derivative over the weight $w$ can be written as follows (by the chain rule):

\begin{equation}
\frac{\partial \mathcal{L}}{\partial w} 
= \frac{\partial \mathcal{L}}{\partial y}\Bigr|_{\hat{y}}\frac{\partial y}{\partial w},
\label{eq:chain}
\end{equation}

where $\hat{y}$ is the (``real'') numerical value of the output $y$. We can thus obtain our desired Hebbian updates as the gradients of the following expressions for $\mathcal{L}$ and $y$:

\begin{gather}
\mathcal{L} = \frac{1}{2}y^2\\
y = \begin{cases} wx  & \mbox{(Plain Hebb)}\\   wx - w^2/2  & \mbox{(Instar rule)} \\ wx - \hat{y}w^2/2  &\mbox{(Oja's rule)} \end{cases} 
\end{gather} 

By using Equation \ref{eq:chain}, it is easily verified that the derivatives of the above expressions over $w$ are $xy$, $y(x-w)$ and $y(x-\hat{y}w)$, respectively. We have also verified that the resulting gradients computed by the framework are equal to hand-computed Hebbian updates in the code.

\lstset{
        language=Python,
        basicstyle=\ttfamily,
        commentstyle=\color{gray},  % requires the xcolor package !
        basewidth = {.5em}  % for some reason, the default listing style has absurdly large space b/w letters
}

The following code implements the method in PyTorch:

\begin{lstlisting}

optimizer.zero_grad()

# Perform the weight convolution, setting up the first part
# of the computational graph.
prelimy = F.conv2d(x, w)

# We now compute the "real" output realy, with a k-WTA
realy = prelimy.clone().detach()  # We don't want to affect the graph
tk = torch.topk(realy.data, K, dim=1, largest=True)[0]
realy.data[realy.data < tk.data[:,-1,:,:][:, None, :, :]] = 0
realy.data = (realy.data > 0).float()  # Binarizing

# We now compute the surrogate output y, used only to set up the
# proper computational graph.
yforgrad = prelimy      # Plain Hebb, dw ~= xy

# The following expressions implement the Instar rule (dw ~= y(x-w)) and Oja's
# rule (dw ~= y(x-wy)), respectively.  Note the dimensional rearrangements and
# broadcasting.
# yforgrad = prelimy - 1/2 * torch.sum(w * w, dim=(1,2,3))[None,:, None, None]
# yforgrad = prelimy - 1/2 * torch.sum(w * w, dim=(1,2,3))[None,:, None, None] * realy.data


 # We overwrite the values of yforgrad with the "real" y.
yforgrad.data = realy.data

# Compute the loss and perform the backward pass, which applies the 
# desired Hebbian updates.
loss =  torch.sum( -1/2 * yforgrad * yforgrad)
loss.backward()
optimizer.step()
\end{lstlisting}

\section{Full sets of learned receptive fields}

\begin{figure}
    \centering
    \includegraphics[scale=.25]{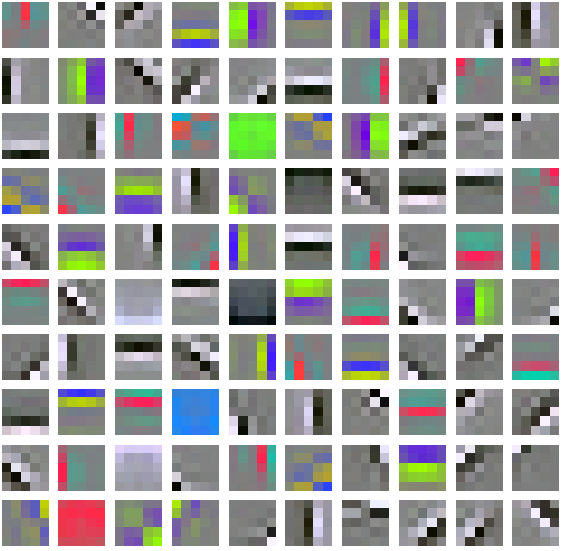}
    \hspace{1pc}
    \includegraphics[scale=.25]{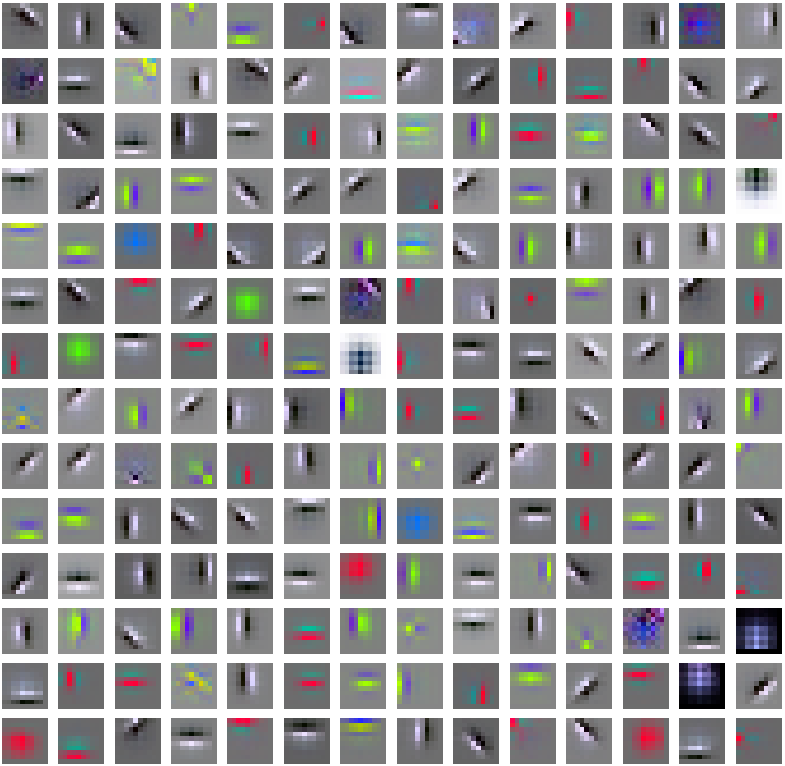}
    \includegraphics[scale=.25]{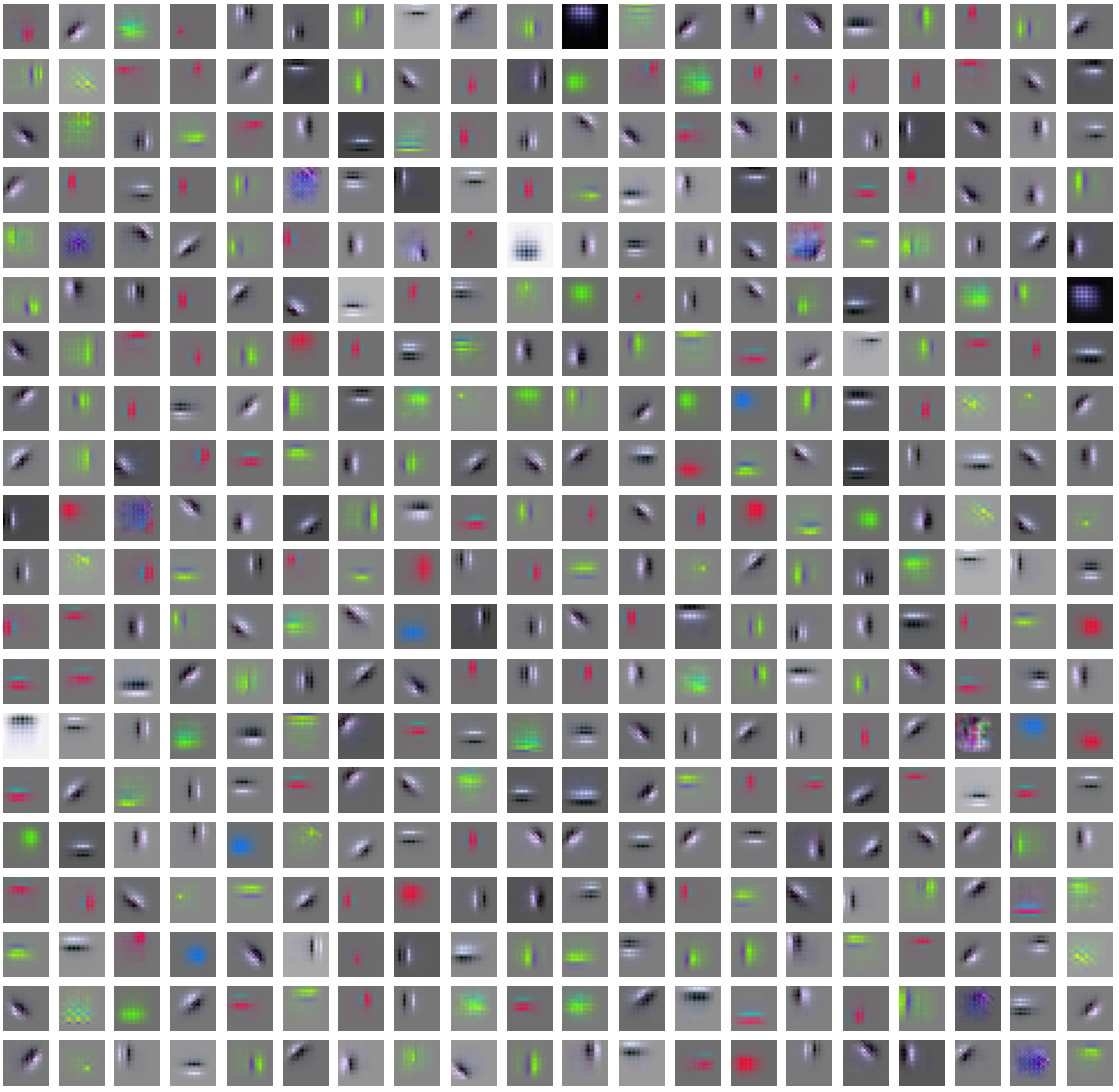}
    \caption{Full set of learned receptive fields from layers 1, 2 and 3 in a standard Hebbian network with WTA competition. }
    \label{fig:RFs_pm0_full}  
\end{figure}

\begin{figure}
    \centering
    \includegraphics[scale=.25]{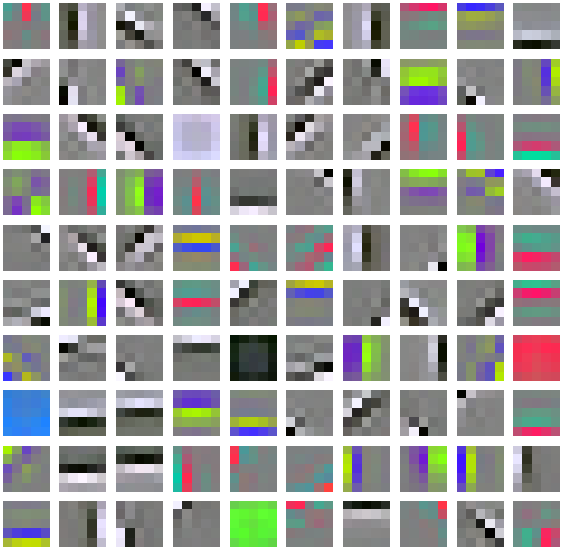}
    \hspace{1pc}
    \includegraphics[scale=.25]{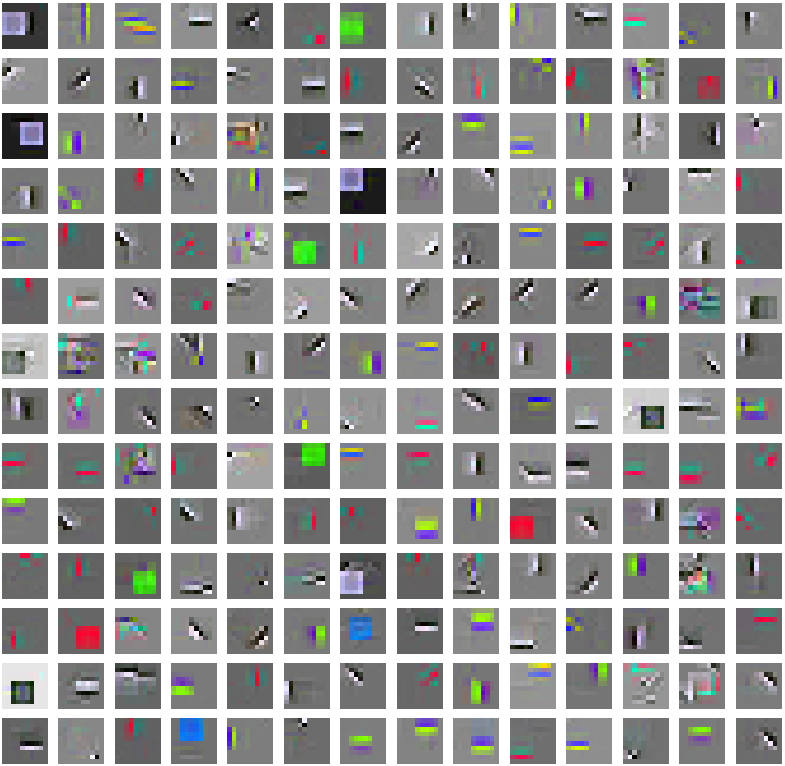}
    %\hspace{1pc}
    \includegraphics[scale=.25]{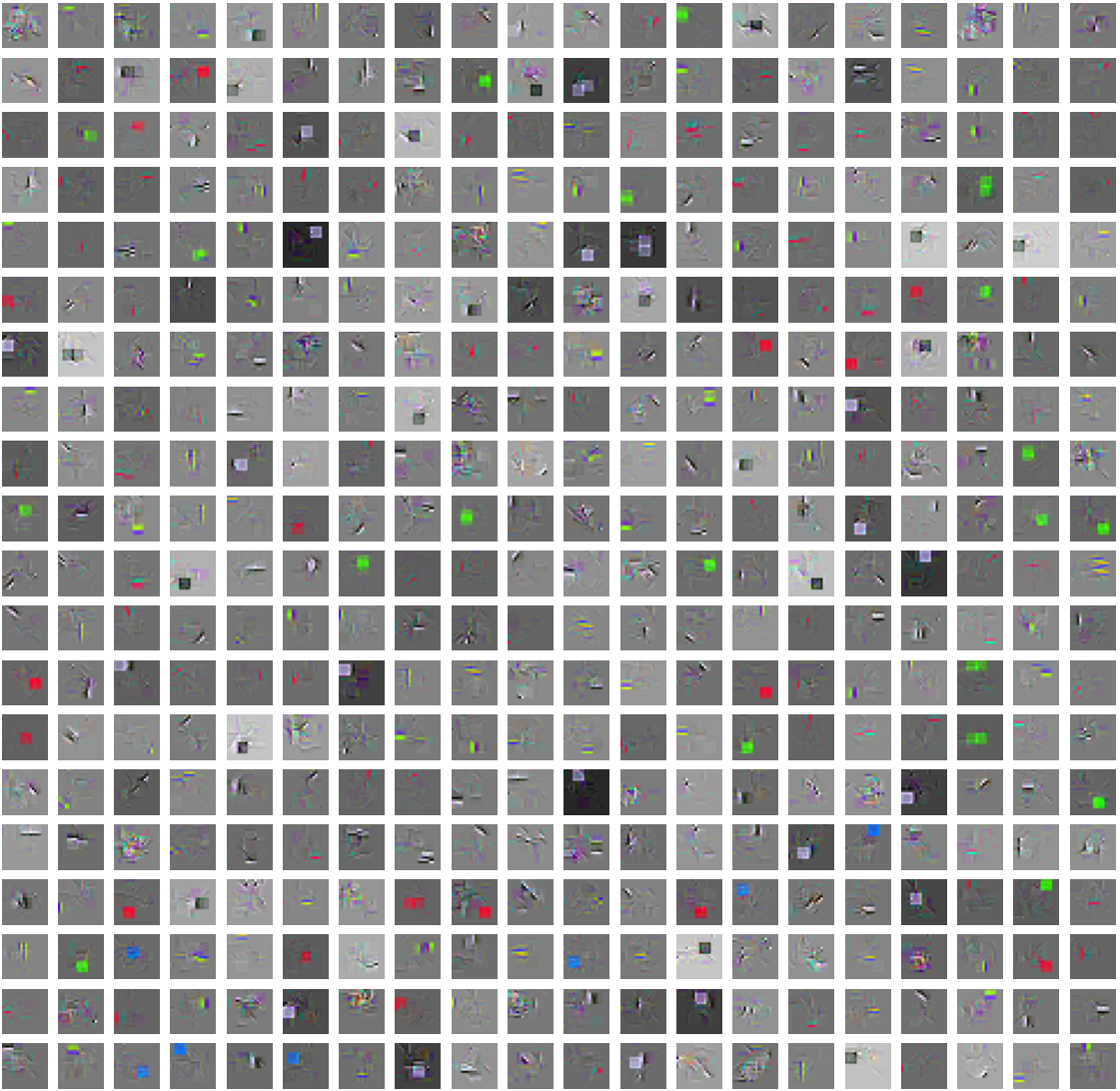}
    \caption{Full set of learned receptive fields from layers 1, 2 and 3 in a Hebbian network with dense activations and pruned connections (plasticity still occurs through WTA competition). }
    \label{fig:RFs_standard_full}  
\end{figure}

Figure \ref{fig:RFs_pm0_full} shows the full set of learned receptive field in the original, unconstrained Hebbian network with WTA competition. Figure \ref{fig:RFs_standard_full} shows the full set of learned receptive field in a  Hebbian network with ``triangle'' activations and pruned connectivity (with plasticity still determined by WTA competition). 

\bibliographystyle{unsrtnat}
\bibliography{references}  %%% Uncomment this line and comment out the ``thebibliography'' section below to use the external .bib file (using bibtex) .

\end{document}